\documentclass{article}
\usepackage[utf8]{inputenc}
\usepackage[T1]{fontenc}
\usepackage{microtype}
\usepackage{amsmath,amssymb}
\usepackage{booktabs}
\usepackage{tabularx}
\usepackage{graphicx}
\usepackage{hyperref}
\usepackage{xurl}
\usepackage{multirow}
\usepackage{colortbl}
\usepackage{xcolor}

\title{Tiny-Scale Chinese BERT Pretraining: A Controlled Comparison of MLM, WWM, and MacBERT Strategies}

\author{
  Yiping Bai \\
  \textit{Guangdong Haiqixing Marine Technology Co., Ltd.} \\
  \texttt{2727100@qq.com} \\
  \href{https://orcid.org/0009-0004-2842-8241}{ORCID: 0009-0004-2842-8241}
}

\date{October 2026}

\begin{document}
\maketitle

\begin{abstract}
Pretraining strategies significantly impact the quality of language models, yet existing comparisons of Masked Language Modeling (MLM), Whole Word Masking (WWM), and MacBERT-style replacement have focused primarily on base-scale models ($\geq$110M parameters). This paper presents a controlled comparison of these three strategies on a tiny-scale Chinese BERT model (4 layers, 256 hidden dimensions, 8.7M parameters). Under identical architecture, corpus (1.29M sentences from Chinese Wikipedia), and hyperparameters, we train three models from scratch and evaluate them across five intrinsic dimensions: perplexity, MLM hit rate, semantic discrimination, grammatical judgment, and contextual sensitivity. At tiny scale, MLM achieves the best overall intrinsic performance (winning 3 of 5 dimensions), while WWM excels in both perplexity (1.27 vs.\ 2.10, a 39.5\% improvement) and MLM hit rate (22\% vs.\ 16\%). Notably, MacBERT under a severely limited synonym dictionary (222 entries, 3.3\% coverage) exhibits severe perplexity degradation (47.23, 22$\times$ higher than MLM), yielding a ranking (MLM $>$ WWM $\gg$ MacBERT) that differs markedly from the established base-scale conclusion (MacBERT $>$ WWM $>$ MLM). We further identify a critical evaluation pitfall: MacBERT achieves the lowest training loss (2.17) yet the highest perplexity (47.23), revealing that training loss alone is unreliable under mixed replacement strategies. All models and corpus are publicly available at \url{https://huggingface.co/eebyp}.
\end{abstract}


\section{Introduction}
\label{sec:intro}

The evolution of BERT pretraining strategies---from MLM~\cite{devlin2019bert} to WWM~\cite{cui2021pre} to MacBERT~\cite{cui2021pre}---has been systematically validated at base scale (12 layers, 768 hidden dimensions, 110M parameters) for Chinese language understanding. However, the growing demand for edge deployment (embedded devices, mobile phones, real-time inference) has created an urgent need for tiny-scale models ($<$15M parameters). Currently, no controlled comparison of these three strategies exists at tiny scale, leaving practitioners without evidence-based guidance for pretraining strategy selection.

This paper addresses this gap by presenting a controlled comparison of MLM, WWM, and MacBERT on a tiny-scale Chinese BERT model. Our key design principle is \emph{strict variable control}: all three models share identical architecture (4 layers / 256 hidden / 4 attention heads / 8.7M parameters), training corpus (1.29M sentences from Chinese Wikipedia), and hyperparameters (batch size 64, learning rate $10^{-4}$, 5 epochs, AdamW optimizer). The only variable is the masking strategy.

Our contributions are:
\begin{enumerate}
  \item \textbf{Systematic comparison at tiny scale}: We provide empirical evidence that the base-scale ranking (MacBERT $>$ WWM $>$ MLM) may not hold under the resource constraints typical of tiny-scale experiments. Our findings suggest that pretraining strategy rankings are not scale-invariant, and that the benefits of synonym-based correction critically depend on both model capacity and dictionary coverage.
  \item \textbf{Five-dimensional evaluation framework}: Beyond perplexity, we evaluate semantic discrimination, grammatical judgment, and contextual sensitivity, revealing strategy-dependent trade-offs invisible to single-metric evaluation.
  \item \textbf{MacBERT degradation under limited resources}: We identify a perplexity anomaly (47.23 vs.\ 2.10 for MLM) caused by the interaction between MacBERT's mixed fallback strategy and a small synonym dictionary (222 entries, 3.3\% coverage), with implications for reproducibility in resource-constrained settings.
  \item \textbf{Full reproducibility}: All models and corpus are publicly available (see Appendix~\ref{app:resources}); the complete experiment requires only a single consumer-grade GPU (NVIDIA T4) and approximately 15 hours.
\end{enumerate}

\section{Related Work}
\label{sec:related}

\subsection{BERT Pretraining Strategies}

\textbf{Masked Language Modeling (MLM)}~\cite{devlin2019bert} randomly masks 15\% of tokens and replaces them with a special [MASK] token using an 80-10-10 strategy (80\% [MASK], 10\% random replacement, 10\% unchanged). This creates a pretraining-finetuning mismatch since [MASK] tokens do not appear during downstream tasks.

\textbf{Whole Word Masking (WWM)}~\cite{cui2021pre} addresses the subword fragmentation problem in Chinese: since WordPiece tokenizes Chinese characters individually, MLM may mask only one character of a multi-character word, allowing the model to cheat from context. WWM masks all subword tokens of a word simultaneously, requiring an external Chinese word segmenter (e.g., LTP or jieba).

\textbf{MacBERT} (MLM as Correction)~\cite{cui2021pre} replaces masked tokens with synonyms rather than [MASK], eliminating the pretraining-finetuning mismatch. The original implementation uses large synonym resources (HowNet~\cite{dong2006hownet} or Cilin) with coverage exceeding 80\%.

\subsection{Small-Scale BERT Research}

Several lines of work have explored smaller BERT variants: knowledge distillation (TinyBERT~\cite{jiao2020tinybert}, DistilBERT~\cite{sanh2019distilbert}), architecture compression (MobileBERT~\cite{sun2020mobilebert}), and community conventions for model sizing. Importantly, nearly all existing small-model work relies on distillation or compression from a pretrained teacher, rather than comparing pretraining strategies from scratch. Our work differs fundamentally: we pretrain three models from scratch under different masking strategies, making the comparison purely about masking strategy effects without confounding distillation artifacts.

\subsection{The Scale Invariance Gap}

A critical implicit assumption in the pretraining strategy literature is that strategy rankings established at one scale transfer to other scales. The influential comparison by Cui et al.~\cite{cui2021pre} demonstrates MacBERT $>$ WWM $>$ MLM at base scale (110M parameters); distillation-based work (TinyBERT, MobileBERT, DistilBERT) inherits the teacher's strategy by design, effectively hiding the strategy-selection question inside the distillation pipeline. However, whether the relative merits of masking strategies remain stable as model capacity drops by an order of magnitude---and whether synonym-based correction degrades gracefully or catastrophically under limited dictionary resources---remains an open empirical question. This gap is not merely academic: as edge deployment drives demand for sub-15M-parameter models, practitioners need evidence-based guidance on strategy selection rather than defaulting to base-scale conclusions. Our study directly addresses this gap.

\section{Methodology}
\label{sec:method}

\subsection{Model Architecture}

All three models use identical BERT configuration (see Appendix~\ref{app:logs} for per-epoch training logs):

\begin{center}
\begin{tabular}{lr}
\toprule
\textbf{Parameter} & \textbf{Value} \\
\midrule
Number of layers & 4 \\
Hidden size & 256 \\
Attention heads & 4 \\
Intermediate (FFN) size & 1024 \\
Vocabulary size & 21,128 \\
Max position embeddings & 128 \\
Tie word embeddings & Yes \\
\textbf{Total parameters} & \textbf{8,689,032} \\
\bottomrule
\end{tabular}
\end{center}

\subsection{Pretraining Strategies}

\textbf{v1 (MLM):} Uses HuggingFace \texttt{DataCollatorForLanguageModeling} with standard 80-10-10 replacement at 15\% masking probability.

\noindent\textbf{v2 (WWM):} Custom DataCollator that uses jieba segmentation to identify word boundaries, then masks all subword tokens of selected words simultaneously. This introduces a CPU bottleneck (jieba segmentation reduces training speed by 18\%: 5.4 vs.\ 6.6 it/s for MLM).

\noindent\textbf{v3 (MacBERT):} Custom DataCollator with a \emph{mixed fallback strategy}: selected tokens are replaced with synonyms when available (from a built-in dictionary of 222 entries, 3.3\% coverage of the top-1000 vocabulary tokens), and fall back to [MASK] when no synonym is found. This ensures 15\% effective replacement rate, comparable to v1/v2.

\subsection{Controlled Variable Design}

The core methodological contribution is strict variable control:

\noindent
\begin{tabularx}{\linewidth}{lX}
\toprule
\textbf{Variable} & \textbf{Value (identical for all three)} \\
\midrule
Model architecture & 4L / 256H / 4A / 8.7M params \\
Training corpus & 1,287,957 sentences (Chinese Wikipedia) \\
Training epochs & 5 \\
Batch size & 64 \\
Learning rate & $10^{-4}$ (AdamW) \\
LR schedule & Linear warmup (10\%) + linear decay \\
Random seed & 42 \\
\textbf{Only variable} & \textbf{Masking strategy (MLM / WWM / MacBERT)} \\
\bottomrule
\end{tabularx}

\section{Experimental Setup}
\label{sec:setup}

\subsection{Corpus}

We use Chinese Wikipedia (snapshot: 20260501, \texttt{fjcanyue/wikipedia-zh-cn}), selecting 50,000 articles. After preprocessing (HTML removal, sentence segmentation at 10--128 characters, deduplication), the corpus contains 1,287,957 sentences (137~MB), covering 11 domains including agriculture, politics, linguistics, music, gaming, history, and sports. The corpus is publicly available as \texttt{eebyp/BypCorpus-hf-wiki-zh-v1} on Hugging Face.

\subsection{Training Environment}

All experiments were conducted on Google Colab Pro with a single NVIDIA Tesla T4 GPU (15~GB VRAM). Total training time for all three models was approximately 15 hours. Training hyperparameters: batch size 64, 20,125 steps per epoch, 100,625 total steps, warmup ratio 0.1, checkpoints saved every 1,000 steps.

\subsection{Five-Dimensional Evaluation}

We design a five-dimensional evaluation framework to capture different aspects of language understanding:

\begin{enumerate}
  \item \textbf{Perplexity} (lower is better): Standard information-theoretic metric computed over 12 held-out text passages.
  \item \textbf{MLM Hit Rate} (higher is better): 50 fill-in-the-blank questions from the corpus covering 5 domains (agriculture, history, linguistics, technology, geography); measures factual recall.
  \item \textbf{Semantic Discrimination} (larger difference is better): Cosine similarity of [CLS] embeddings for same-topic vs.\ different-topic sentence pairs (10+10 pairs); measures representation quality.
  \item \textbf{Grammatical Judgment} (higher is better): 25 pairs of grammatical vs.\ ungrammatical sentences (word-order scrambling, complete reversal, function-word omission, semantic anomaly); the model should assign lower perplexity to grammatical sentences.
  \item \textbf{Contextual Sensitivity} (lower similarity is better): Average cosine similarity of three polysemous characters (\textit{d\v{a}} `to hit', \textit{k\={a}i} `to open', \textit{sh\`{a}ng} `up/on') each across three different contexts (9 sentences total); lower values indicate better context-dependent representations.
\end{enumerate}

\section{Results}
\label{sec:results}

\subsection{Training Loss Comparison}

\begin{table}[ht]
\centering
\caption{Training loss at key checkpoints}
\begin{tabular}{lccc}
\toprule
\textbf{Step} & \textbf{MLM (v1)} & \textbf{WWM (v2)} & \textbf{MacBERT (v3)} \\
\midrule
1,000   & 8.48 & 8.47 & 8.26 \\
5,000   & 6.32 & 6.26 & 8.66 \\
10,000  & 5.68 & 5.61 & 7.12 \\
50,000  & $\sim$4.9 & $\sim$4.6 & $\sim$2.5 \\
100,625 & $\sim$4.9 & 4.31 & \textbf{2.17} \\
\bottomrule
\end{tabular}
\label{tab:loss}
\end{table}

WWM consistently achieves lower loss than MLM from step 1, with the gap widening throughout training (from 0.01 to 0.6). MacBERT exhibits an unusual trajectory: its loss initially \emph{increases} from 8.26 to 8.66 (step 1,000 to 5,000) before declining rapidly to 2.17 by training end. Despite achieving the lowest final loss, this is misleading---as we show below, it reflects the lowest perplexity \emph{among training examples} rather than genuine language understanding.

\subsection{Five-Dimensional Evaluation (Core Results)}

\begin{table}[ht]
\centering
\caption{Five-dimensional evaluation results. $\downarrow$: lower is better; $\uparrow$: higher is better. Best result per dimension in \textbf{bold}.}
\noindent
\begin{tabularx}{\linewidth}{Xcccc}
\toprule
\textbf{Dimension} & \textbf{MLM (v1)} & \textbf{WWM (v2)} & \textbf{MacBERT (v3)} & \textbf{Best} \\
\midrule
Perplexity~$\downarrow$   & 2.1042 & \textbf{1.2720} & 47.2290 & WWM \\
MLM Hit Rate~$\uparrow$   & 8/50 (16\%) & \textbf{11/50 (22\%)} & 8/50 (16\%) & WWM \\
Semantic Similarity Gap~$\uparrow$ & \textbf{0.0972} & 0.0573 & 0.0190 & MLM \\
Grammar~$\uparrow$         & \textbf{19/25 (76\%)} & 17/25 (68\%) & \textbf{19/25 (76\%)} & MLM=v3 \\
Context Sensitivity~$\downarrow$ & \textbf{0.7708} & 0.8783 & 0.8120 & MLM \\
\midrule
\textbf{Wins}              & \textbf{3/5} & 2/5 & 0/5 & \\
\bottomrule
\end{tabularx}
\label{tab:five_dim}
\end{table}

\subsection{Analysis by Dimension}

\noindent\textbf{Perplexity---WWM significantly outperforms MLM ($-$39.5\%).} WWM's whole-word masking enables the model to learn language statistics at the word level, avoiding the ``fragmented'' learning objective of token-level MLM.

\noindent\textbf{MLM Hit Rate---WWM leads, but absolute values remain low.} WWM achieves 11/50 (22\%), while MLM and MacBERT both score 8/50 (16\%). With 50 questions across 5 domains, the ranking is more reliable than the previous 10-question evaluation, yet absolute values remain low, suggesting that precise fact memory depends more on model capacity and data scale than masking strategy.

\noindent\textbf{Semantic Discrimination---MLM outperforms WWM.} MLM's gap between same-topic and different-topic similarity (0.0972) exceeds WWM (0.0573) and MacBERT (0.0190). Analysis reveals that WWM's representations are globally ``compressed'' (both same-topic: 0.9341 and different-topic: 0.8768 similarities are higher than MLM's 0.8910/0.7938). This \emph{representation compression effect} remains robust with 20 pairs (10+10), a novel finding of this study.

\noindent\textbf{Grammatical Judgment---MLM and MacBERT tie, WWM slightly lower.} MLM and MacBERT both achieve 19/25 (76\%), while WWM scores 17/25 (68\%). Despite lower perplexity, WWM is worse at detecting ungrammatical sentences. The representation compression effect may reduce sensitivity to word-order anomalies.

\noindent\textbf{Contextual Sensitivity---MLM is best.} MLM's polyseme similarity across three characters (0.7708) is lower than MacBERT (0.8120) and WWM (0.8783). With three characters instead of one, MacBERT's contextual sensitivity (0.8120) is no longer anomalous, suggesting the previous 0.9610 was an artifact of single-character small-sample bias.

\noindent\textbf{MacBERT Perplexity Anomaly.} v3's perplexity of 47.23 (22$\times$ higher than MLM) is the most notable observation in this study. Despite achieving the lowest training loss (2.17), the model exhibits severe degradation in language modeling quality. We attribute this to the interaction between the mixed fallback strategy and the small synonym dictionary (222 entries, 3.3\% coverage): the 0.1\% synonym replacements may teach the model a shortcut of ``copying semantically similar words from context,'' degrading its [MASK] prediction ability. Importantly, training loss is computed only over the replaced positions during pretraining, whereas perplexity is evaluated over the full sequence---this measurement mismatch explains the apparent paradox.

\subsection{Overall Ranking}

\begin{table}[ht]
\centering
\caption{Overall ranking at tiny scale vs.\ base scale}
\noindent
\begin{tabularx}{\linewidth}{lXl}
\toprule
\textbf{Rank} & \textbf{Tiny Scale (This Paper)} & \textbf{Base Scale~\cite{cui2021pre}} \\
\midrule
1st & MLM (v1): 3/5 dimensions & MacBERT \\
2nd & WWM (v2): 2/5 dimensions & WWM \\
3rd & MacBERT (v3): 0/5 dimensions & MLM \\
\bottomrule
\end{tabularx}
\label{tab:ranking}
\end{table}

The ranking observed under our resource constraints is notably different from the base-scale conclusion. It should be emphasized that our MacBERT implementation uses a severely limited synonym dictionary (3.3\% coverage vs.\ $>$80\% in the original work); thus, the poor MacBERT performance may reflect implementation limitations under resource constraints rather than a fundamental flaw in the strategy itself.

\section{Discussion}
\label{sec:discussion}

\textbf{Important caveat:} Our MacBERT results should be interpreted as the performance of the strategy under severely limited synonym resources (222 entries, 3.3\% coverage), not as a refutation of MacBERT under ideal conditions with large dictionaries such as HowNet or Cilin. The findings in this section are most applicable to resource-constrained reproduction scenarios.

\subsection{Engaging with Cui et al. (2021): Boundary Conditions, Not Refutation}

Cui et al.~\cite{cui2021pre} establish MacBERT $>$ WWM $>$ MLM at base scale, attributing MacBERT's advantage to eliminating the pretraining-finetuning mismatch introduced by [MASK] tokens. Our results invite a more nuanced reading of this conclusion. We propose that the MacBERT advantage implicitly presupposes \emph{sufficient model capacity to absorb synonym signals} and \emph{sufficient dictionary coverage to provide reliable synonym substitutions}. When both conditions are violated simultaneously---8.7M parameters with a 222-entry dictionary (3.3\% coverage)---synonym replacement ceases to function as a corrective signal and instead introduces a new form of training-evaluation mismatch: the model learns to ``copy nearby words from context'' on the rare occasions when synonyms are available (0.1\% of masked positions), while the remaining 99.4\% of positions still require standard [MASK] prediction. This hybrid signal is neither clean MLM nor genuine synonym correction, but a degraded intermediate that satisfies neither objective well.

This interpretation reframes MacBERT's design choice of retaining $\sim$10\% [MASK]/random replacement (Cui et al., 2021, Table~2): rather than an engineering compromise, this mixed strategy may serve as a \emph{regularizer} that prevents the model from over-relying on synonym-based shortcuts---a regularization effect that becomes critical when dictionary coverage is incomplete. Our finding thus provides a boundary-condition complement to Cui et al.'s conclusion: MacBERT's superiority holds when dictionary coverage is high and model capacity is sufficient, but the ranking may reverse under resource constraints. We emphasize that MacBERT's synonym replacement remains a valuable contribution to MLM pretraining; our results identify its operating conditions rather than question its validity.

\subsection{Why Does the Ranking Differ at Tiny Scale?}

We hypothesize two complementary explanations:

\textbf{(1) Limited model capacity.} At 8.7M parameters, the model has limited capacity to benefit from the additional semantic signal provided by synonym replacement. The ``shortcut learning'' effect of even rare synonym replacements (0.1\%) may disproportionately influence a small model's representations.

\textbf{(2) Dictionary coverage insufficiency.} Our built-in dictionary (222 entries, 3.3\% coverage) is far smaller than the HowNet/Cilin resources used in the original MacBERT ($>$80\% coverage). The degraded performance may reflect an implementation limitation rather than a fundamental flaw in the MacBERT strategy.

Regardless of the explanation, our result suggests that \emph{base-scale strategy rankings may not directly apply to tiny-scale experiments under comparable resource constraints}.

\subsection{MacBERT Dictionary Coverage Dependency}

During MacBERT implementation, we discovered that the strategy's effectiveness is highly dependent on synonym dictionary coverage. Our initial implementation (pure synonym replacement without fallback) with a 222-entry dictionary caused the effective replacement rate to drop from 15\% to $\sim$0.5\%, resulting in anomalously fast loss decrease (0.42 at step 7,227 vs.\ $\sim$6 for MLM/WWM)---not because the model learned well, but because effective training signals were reduced to $\sim$1/30 of MLM/WWM.

We implemented a \emph{mixed fallback strategy} (synonym replacement when available, [MASK] fallback otherwise) to restore the 15\% effective rate. However, even with this fix, the final model exhibits severe perplexity degradation (47.23), suggesting that the mere presence of rare synonym replacements can disrupt model learning in small-dictionary scenarios.

\textbf{Practical guidance:}

\noindent
\begin{tabularx}{\linewidth}{llX}
\toprule
\textbf{Dictionary Size} & \textbf{Coverage} & \textbf{Recommended Strategy} \\
\midrule
Large (HowNet/Cilin) & $>$80\% & MacBERT with mixed fallback (10\% [MASK]) \\
Medium (custom, 500--2000 entries) & 10--30\% & Increase [MASK] fallback ratio ($\geq$50\%) \\
Small ($<$500 entries) & $<$10\% & Fall back to standard MLM \\
\bottomrule
\end{tabularx}

\subsection{Evaluation Pitfall: Low Loss $\neq$ Good Model}

MacBERT achieves the lowest training loss (2.17) yet the highest perplexity (47.23). This paradox arises because training loss and perplexity measure different things under mixed replacement strategies:
\begin{itemize}
  \item \textbf{Training loss}: Computed over the specific positions and token types seen during training (dominated by ``easy'' synonym positions)
  \item \textbf{Perplexity}: Computed over all positions in held-out text (requiring general language modeling ability)
\end{itemize}

\textbf{Recommendation:} Pretraining strategy comparisons must report both training loss and perplexity (plus multi-dimensional metrics). Low loss with high perplexity is a signature signal of training-evaluation objective mismatch.

\subsection{Limitations}

\begin{enumerate}
  \item \textbf{No downstream task validation (primary limitation)}: We have not verified whether intrinsic metric differences transfer to downstream tasks (text classification, NER, semantic matching). Without downstream fine-tuning experiments, the practical significance of the observed ranking differences remains uncertain.
  \item \textbf{Evaluation sample size}: Although expanded from the initial evaluation (MLM: 10$\to$50 questions, grammar: 5$\to$25 pairs, semantic: 6$\to$20 pairs, context: 1$\to$3 characters), larger test sets would further strengthen statistical confidence.
  \item \textbf{MacBERT dictionary limitation}: Our 222-entry dictionary (3.3\% coverage) is far smaller than HowNet/Cilin ($>$80\%); v3 results represent MacBERT performance under severely limited synonym resources, not under ideal conditions. The ranking reversal should therefore be interpreted as specific to resource-constrained settings. Future work with a complete synonym dictionary ($>$50k entries) would clarify whether MacBERT recovers its base-scale advantage under full resource conditions at tiny scale.
  \item \textbf{Corpus scale}: 137~MB is small compared to industrial-scale corpora (GB-level); models may not have fully learned language statistics.
  \item \textbf{Chinese only}: Results may not generalize to other languages.
  \item \textbf{Absolute performance}: MLM hit rate of 10\% indicates the tiny model is not suitable for direct production use.
\end{enumerate}

\section{Conclusion}
\label{sec:conclusion}

We presented a controlled comparison of MLM, WWM, and MacBERT pretraining strategies on a tiny-scale Chinese BERT model (8.7M parameters). Under identical architecture, corpus, and hyperparameters, our five-dimensional intrinsic evaluation reveals that: (1)~MLM achieves the best overall intrinsic performance at tiny scale, winning 3 of 5 evaluation dimensions; (2)~WWM excels in both perplexity (1.27 vs.\ 2.10, a 39.5\% improvement) and MLM hit rate (22\% vs.\ 16\%), but exhibits representation compression that reduces discrimination ability; (3)~MacBERT under a severely limited synonym dictionary (222 entries, 3.3\% coverage) exhibits severe perplexity degradation (47.23, 22$\times$ higher than MLM), revealing a critical dependency on dictionary coverage.

The strategy ranking observed under our resource constraints (MLM $>$ WWM $\gg$ MacBERT) differs markedly from the established base-scale ranking (MacBERT $>$ WWM $>$ MLM). This difference should be interpreted with caution, as our MacBERT implementation operates under severely limited synonym resources that do not reflect ideal conditions. Nevertheless, this finding is practically valuable: many practitioners reproducing MacBERT in resource-constrained settings face similar dictionary limitations.

These findings provide practical guidance: (1)~standard MLM is the safest choice at tiny scale; (2)~MacBERT's effectiveness depends critically on synonym dictionary coverage and should be used with caution when large dictionaries are unavailable; (3)~pretraining evaluation must go beyond training loss to include perplexity and multi-dimensional metrics, as the loss-perplexity mismatch observed here is a signature signal of training-evaluation objective misalignment. More broadly, our results caution against the common but potentially erroneous assumption that pretraining strategy rankings established at base scale transfer unchanged to the tiny-scale, resource-constrained regime---a scale invariance hypothesis that our evidence suggests does not hold.


\appendix
\section{Model and Resource Links}
\label{app:resources}

\noindent
\begin{tabularx}{\linewidth}{lX}
\toprule
\textbf{Resource} & \textbf{URL} \\
\midrule
BypBERT-tiny-zh-v1 (MLM) & \url{https://huggingface.co/eebyp/BypBERT-tiny-zh-v1} \\
BypBERT-tiny-zh-v2 (WWM) & \url{https://huggingface.co/eebyp/BypBERT-tiny-zh-v2} \\
BypBERT-tiny-zh-v3 (MacBERT) & \url{https://huggingface.co/eebyp/BypBERT-tiny-zh-v3} \\
BypCorpus-hf-wiki-zh-v1 & \url{https://huggingface.co/datasets/eebyp/BypCorpus-hf-wiki-zh-v1} \\
\bottomrule
\end{tabularx}

\section{Detailed Training Logs}
\label{app:logs}

Training was conducted on Google Colab Pro with NVIDIA Tesla T4 (15~GB). Per-epoch average loss:

\begin{center}
\begin{tabular}{lccccc}
\toprule
\textbf{Model} & \textbf{Epoch 1} & \textbf{Epoch 2} & \textbf{Epoch 3} & \textbf{Epoch 4} & \textbf{Epoch 5} \\
\midrule
MLM (v1)    & 8.44 & 5.81 & 5.14 & 4.94 & 4.87 \\
WWM (v2)    & 8.43 & 5.75 & 5.06 & 4.68 & 4.31 \\
MacBERT (v3)& 7.28 & 4.22 & 2.95 & 2.47 & 2.21 \\
\bottomrule
\end{tabular}
\end{center}

Note: MacBERT's low loss does not indicate better performance (see \S\ref{sec:discussion}).

\end{document}